\documentclass[letterpaper, 10 pt, conference]{ieeeconf}  % Comment this line out if you need a4paper

\IEEEoverridecommandlockouts                              % This command is only needed if 
\usepackage{epsfig} % for postscript graphics files
\usepackage{epstopdf}
\usepackage{graphicx}
\usepackage{graphics}
\usepackage{color}
\usepackage{soul}

\title{\LARGE \bf
A Humanoid Social Agent Embodying Physical Assistance Enhances Motor Training Experience
}

\IEEEaftertitletext{ \copyright 2020 IEEE. Personal use of this material is permitted. Permission from IEEE must be
obtained for all other uses, in any current or future media, including reprinting/republishing this material for advertising or promotional purposes, creating new collective works, for resale or redistribution to servers or lists, or reuse of any copyrighted component of this work in other works. 
DOI: 10.1109/RO-MAN47096.2020.9223335}

\author{Giulia Belgiovine$^{1,2}$, Francesco Rea$^{1}$, Jacopo Zenzeri$^{1}$ and Alessandra Sciutti$^{3}$% <-this % stops a space
\thanks{*This work was supported by a Starting Grant from the European Research Council (ERC) under the European Union’s Horizon 2020 research and innovation programme. G.A. No 804388, wHiSPER}% <-this % stops a space
\thanks{$^{1}$GB ({\tt\small giulia.belgiovine@iit.it}), FR and JZ are with the Robotics, Brain and Cognitive Sciences Unit, Istituto Italiano di Tecnologia, Genova, Italy
}%
\thanks{$^{2}$GB is with Department of Informatics, Bioengineering, Robotics,and System Engineering, University of Genoa, Genoa, Italy
        }%
\thanks{$^{3}$AS is with the CONTACT unit, Istituto Italiano di Tecnologia, Genova, Italy}}

\begin{document}

\graphicspath{ {./figure/} }

\maketitle
\thispagestyle{empty}
\pagestyle{empty}

%%%%%%%%%%%%%%%%%%%%%%%%%%%%%%%%%%%%%%%%%%%%%%%%%%%%%%%%%%%%%%%%%%%%%%%%%%%%%%%%
\begin{abstract}

Skilled motor behavior is critical in many human daily life activities and professions. The design of robots that can effectively teach motor skills is an important challenge in the robotics field. In particular, it is important to understand whether the involvement in the training of a robot exhibiting social behaviors impacts on the learning and the experience of the human pupils. In this study, we addressed this question and we asked participants to learn a complex task - stabilizing an inverted pendulum - by training with physical assistance provided by a robotic manipulandum, the Wristbot. One group of participants performed the training only using the Wristbot, whereas for another group the same physical assistance was attributed to the humanoid robot iCub, who played the role of an expert trainer and exhibited also some social behaviors. The results obtained show that participants of both groups effectively acquired the skill by leveraging the physical assistance, as they significantly improved their stabilization performance even when the assistance was removed. Moreover, learning in a context of interaction with a humanoid robot assistant led subjects to increased motivation and more enjoyable training experience, without negative effects on attention and perceived effort. With the experimental approach presented in this study, it is possible to investigate the relative contribution of haptic and social signals in the context of motor learning mediated by human-robot interaction, with the aim of developing effective robot trainers.

\end{abstract}

%%%%%%%%%%%%%%%%%%%%%%%%%%%%%%%%%%%%%%%%%%%%%%%%%%%%%%%%%%%%%%%%%%%%%%%%%%%%%%%%
\section{INTRODUCTION}

Skilled motor behavior is critical in many human daily life activities and professions; additionally, it represents the basis for many of the highest human endeavors and cultural achievements, like sport, art, and music.

The transfer of skills from an expert to a naive person is an example of a social complex interaction involving different facets of human communication. In most cases, the knowledge transferred from the expert to the naive is explicit, but often it largely transfers in implicit forms, relying on both contact and non-contact exchanged messages.
This is what happens for example when a therapist trains a patient to recover certain motions; during such interaction, the two actors exchange several mutual information through different communication channels, that enables the therapist to gain knowledge not only about the patient's physical improvements but also about his/her cognitive and emotional state. 
Recently, there has been great interest in addressing skill transfer in the context of physical interactions \cite{Mireles_2017}, \cite{Ganesh_2014}. 
One of the crucial limits is that measuring the multimodal components of interaction (from the physical to the social ones) is quite complex. Robots can help to address this issue, by enabling to study interaction while it unfolds and ensuring controllability and reproducibility \cite{Sciutti_2017}.

In this context, several studies have focused mainly on physical interactions between humans and robots to define the optimal strategy to train humans \cite{Galofaro_2017}, \cite{Zenzeri_2014}.
Other researchers have instead considered the social aspects that robots should present to be effective teachers, tutors, or peer learners (see \cite{Belpaeme_2018} for a recent review). Studies that investigate the relative roles of the physical and the social components in a single, unified setting are more scarce \cite{Gorsic_2017}, \cite{Granados_2017}.

The integration of physical information with social and cognitive cues in the decision-making process of social trainer robots can lead to significant benefits to optimize the learning experience of the human trainees. Indeed, motor learning, and motor behavior in general, has a strong cultural, social, cognitive and affective nature \cite{Wulf_2016} and it cannot be completely understood, and consequently optimized, without considering the influences of all these aspects concomitantly. 

Another important aspect of effective motor learning is the ability to generalize the skills acquired during training.
It is well known that learning a new skill requires the acquisition of an internal model of the dynamics of the task \cite{Wolpert_1998}; by exploring and exploiting this model, learners become proficient in that task and are able to adapt to changes of the working environment.
When the task is facilitated by assistance during training (as is often the case in rehabilitation), the representation created by the naive learner may be distorted by the inclusion of assistance as part of the task dynamic, rather than as a transient contribution that helps to achieve the task. 
The training must, therefore, be aimed at promoting the learner's ability to generalize the acquired model of the task, adapting and exploiting it when the robot's help is no longer present.

Building an appropriate and effective robotic teaching architecture implies facing several challenges as (i) understanding the effects of the presence of a social robot as a trainer on the performance and experience of the subjects; (ii) identifying which are the most significant cues the robot has to leverage in the decision-making process and, finally, (iii) selecting which signals to convey for effective teaching and how to communicate them.

In this work, we addressed the first major challenge and investigated the impact of the introduction of a humanoid robot as an expert assistant in a motor learning task.

Mounting evidence has shown that specialized robots can influence both human performance and motivation during physical and cognitive interactions \cite{Fasola_2010}. For example, the presence of a humanoid robot co-worker has demonstrated to have some influence on human performance frequency, without any effect on performance accuracy \cite{Vasalya_2018}. 
The same study highlights the importance of the physical presence and the appearance of the robot in this context. Indeed, people tend to show stronger behavioral and attitudinal responses to a physically real agent as opposed to a virtual one \cite{Li_2015}. 

With the current experiment, we wanted to verify whether and how interacting with a humanoid robot presenting social behaviors and embodying the physical assistance affected the performance and the experience of naive humans intent in learning a new motor skill. 

To answer our research questions, we asked participants to learn a complex motor task - stabilizing an inverted pendulum - by training with physical assistance provided by a robotic manipulandum, the Wristbot \cite{Masia_2009}, \cite{Iandolo_2019}. In particular, we confronted two groups of participants with two different experimental conditions. One group performed the training with physical assistance provided by the Wristbot (\emph{Control group}) while the other group experienced the same assistance, but the help was attributed to the humanoid robot iCub \cite{iCub}, who exhibited some social behaviors (as gazing and talking) and played the role of the expert assistant (\emph{iCub group}). The comparison of the learning performance and the experience evaluation between the two groups allowed us to quantify the impact of the involvement of the humanoid trainer.

More in detail, we tested the hypothesis that learning in an interaction context with a social humanoid robot promotes engagement and motivation of the naive subjects, leading to a more enjoyable training experience.

Moreover, we evaluated whether, in such a demanding task, where it is essential to remain focused on an external stimulus, the presence of a humanoid partner could lead to negative effects on subjects' concentration and perceived effort, with possible negative consequences on training performance.
 
Last, we assessed whether we could promote participants' ability to generalize the learned skill by explicitly presenting the assistance as the contribution of a second agent during the training. We speculate that this could indeed facilitate distinguishing the transitional contributions of assistance from the intrinsic dynamics of the working environment.

\section{METHODS}

\subsection{Participants}

For this study, 32 participants were recruited in total (14 male, 18 females). We tested 16 subjects in the \emph{iCub group} (7 males, 9 females, 26.1$\pm$3.9 years of age) and the remaining 16 in the \emph{Control group} (7 males, 9 females, 27.1$\pm$3.1 years of age). All participants were right-handed and did not have any known neurological or physical impairment. All participants gave their written informed consent before participating. The research conforms to the ethical standards laid down in the 1964 Declaration of Helsinki, which protects research subjects, and was approved by the local ethical committee of the Liguria Region (n. 222REG2015). 

\subsection{Experimental Setup}

\begin{figure*}%[htp]
  \centering
{\includegraphics[width=2.5in]{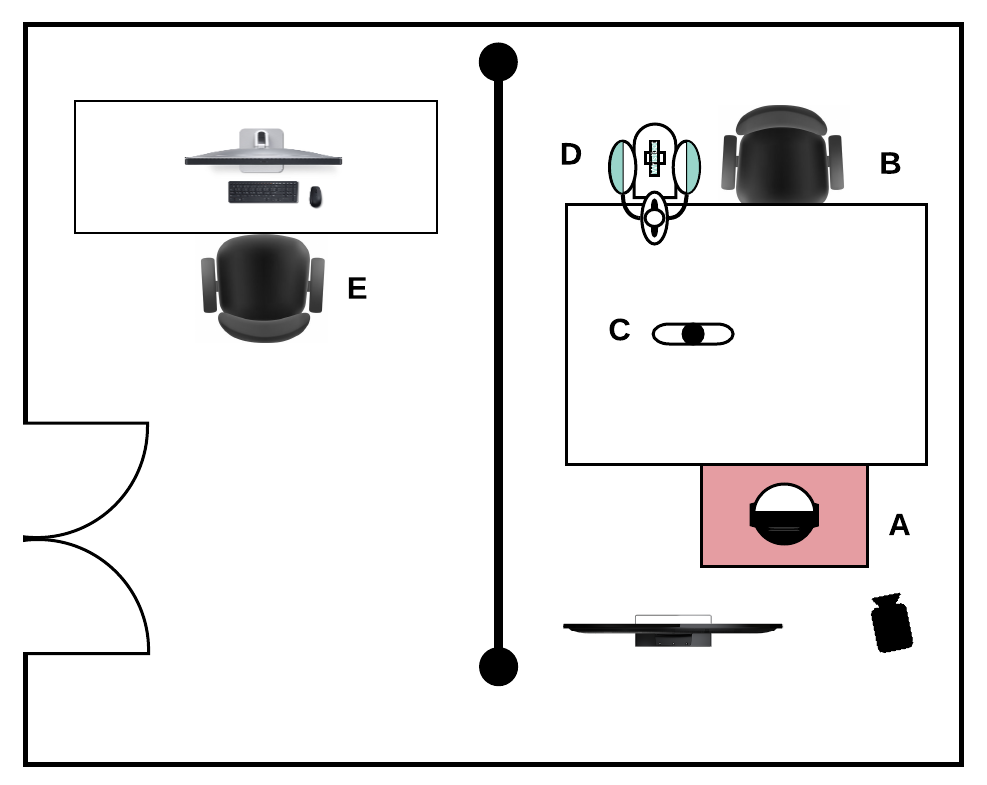}} \quad
{\includegraphics[scale = 0.064]{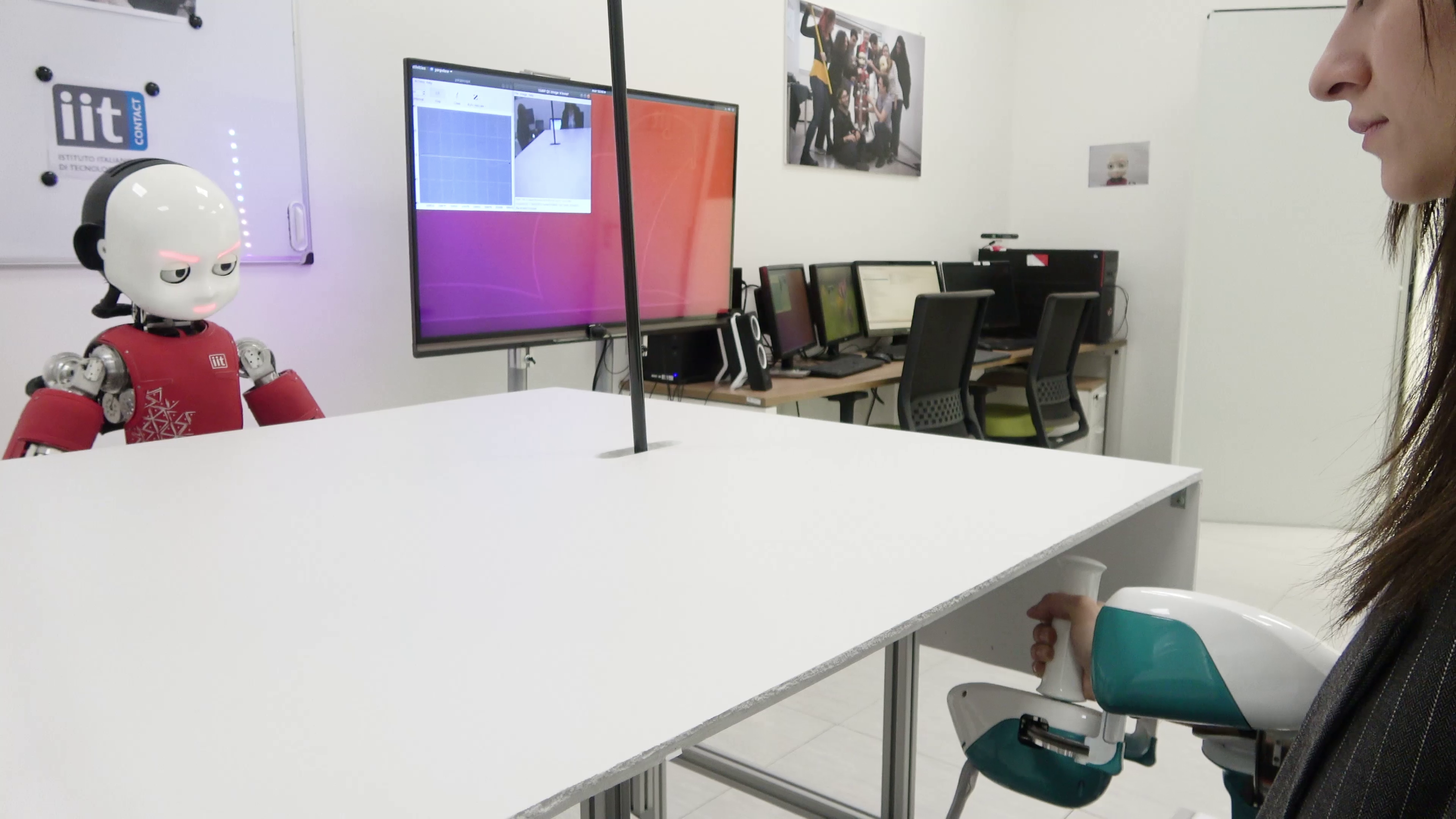}}
  \caption{Left Panel: the setup of the experimental room for the \emph{iCub group}. The iCub robot (A) is standing in front of the participant (B), who is sitting next to the Wristbot (D). The pendulum is between them (C), while the experimenter (E) is hidden behind a black curtain. For the \emph{Control group}, the set-up was the same except for the iCub robot that was hidden behind another black curtain. Right Panel: The robot gazing with a focused expression at the participant, who is trying to balance the pendulum using the Wristbot.}
\label{Setup}
\end{figure*}

During the experiment, participants were sitting on a fixed chair, facing a support (of the height of 76 cm) on which an inverted pendulum structure was mounted. The pendulum was constituted of a 52 cm long rod, made of carbon fiber, and linked on its basis to a brushless motor. The maximal angular excursion of the pendulum was equal to $\pm$40 degrees.
On the right side of the chair, it was placed a robotic manipulandum, the Wristbot, that served as a haptic joystick to deliver forces directly to the inverted pendulum and control its position. The Wristbot permitted, in this specific case, only movements on the Pronation/Supination (PS) plane of the human wrist, with a maximal angular excursion of $\pm$60 degrees.
For the \emph{iCub group}, the humanoid robot iCub was positioned on the other side of the pendulum structure, with the torso at 60 cm from it and on a fixed platform. A squared surface (1 m x 1 m) covered the whole structure and the hands of both the subject and the robot (Fig. \ref{Setup}).
After explaining the task to the participants, the experimenter sat behind the curtain hiding the workstation and started the execution of the routines, controlling the transition between different robot's states. Except for the face-tracking behavior, the robot was not responsive to stimuli from participants.
For the \emph{Control group}, the iCub was hidden behind a black curtain.

%(foto e schema del setup)

\subsection{Experimental Protocol and Task}

For both the experimental groups, the protocol comprised 3 different sessions: i. baseline (1 trial), ii. training (5 trials) and iii. test (3 trials).
Each trial lasted 2.5 minutes: during this period subjects were asked to accomplish the stabilization task, namely to maintain the unstable inverted pendulum in balance, by controlling its position through the angular orientation of the Wristbot. Their goal was to keep this balancing condition as long as possible. Between each trial, a pause of 2.5 minutes was due, to prevent subjects' fatigue.

Moreover, auditory feedback was provided such that when the pendulum was in balance (i.e., in a range of $\pm$5 degrees from the vertical) no sound was heard; as soon as the pendulum rod left its balance region, a tone was emitted (the frequency of which changed depending on the direction in which the pendulum fell) with a magnitude that increased with the distance from the vertical.

Before the experiment began, the experimenter welcomed the participant and provided a brief explanation about how the experiment was structured, what the task consisted of, and what the final goal was. The experimenter explained explicitly to the subjects that during the training phase there was assistance coming from the Wristbot or iCub, depending on the experimental condition.

For the \emph{iCub group}, also the robot itself, before starting the baseline session, gave a brief introductory explanation about the objective of the task and its role as an assistant during the training phase.
Furthermore, the experimenter clarified that during the resting phases it was not possible to speak with iCub and that, in any case, iCub was not allowed to respond to any interaction.

To make the task challenging and not-trivial, we implemented virtual dynamics that determined the angular orientation of the pendulum on the basis of the angular orientation of the Wristbot handle. The dynamics included a non-linear spring, which virtually connected the Wristbot to the pendulum, and an unstable viscous force-field in which the pendulum was immersed.
%and that increased the speed of the pendulum as it fell.
The designed task required participants to monitor and act on the position of the pendulum almost continuously (through both visual and audible feedback).

The parameters characterizing the system dynamics, as well as trial duration and the average number of trials needed to learn the successful strategy, were chosen after a previous pilot study conducted on a sample population similar to the one of this study. Our goal was to have a task that met an optimal challenge level, without resulting too easy and leading to boredom, or too arduous to allow for any learning to occur.

During the training phase, the task dynamics was facilitated by assistance that reduced the instability of the viscous force-field of 30\%.
The assistance level, selected after piloting, was designed not to fully compensate for the instability, but rather to facilitate the task while maintaining it still challenging, to enable the learning process.
The implemented assistance wanted to emulate the help of a trainer that intervenes in the task by dampening the fall of the pendulum, making it easier to control. 

The assistance did not change during the whole training session and did not adapt to participants' improvements. This choice met the need to keep the experimental conditions as comparable as possible between the \emph{iCub group} and the \emph{Control group}. 
On the other hand, this type of sub-optimal assistance made it necessary to consider as exclusion criteria the removal of those subjects for whom the task was too difficult and non-feasible to learn under the chosen conditions.
Therefore, we considered as outliers the participants who were unable to learn in this context, i.e. who exhibited improvements in performance after training of less than about 1\%, corresponding to the mean minus 1.5 times the standard deviation of the whole population. According to this criterion, we excluded from further analysis one subject from the \emph{Control group} and 2 subjects from the \emph{iCub group}.
In the test phase, the task was the same as the one faced during baseline, with no assistance.

\subsection{Interaction with the Robot iCub}

In this experiment, iCub acted as the expert assistant facilitating the task in the training phase. To this aim, iCub performed specific prono-supination movements of its forearm mimicking the control of the motorized pendulum. The staging was effective, in fact when questioned after the experiment, all participants did not doubt that the assistance came from the intervention of the humanoid robot.

In order to investigate the effect of the involvement of the humanoid robot on subjects' experience, we wanted to keep the two experimental conditions as comparable as possible. Hence, iCub's social behaviors did not adapt to participants' performance, to emulate the condition in the \emph{Control group}, where no feedback was provided by the Wristbot.

The following describes iCub's behavior during the interaction. At the beginning of the experiment, the robot welcomed the participant, gave a brief explanation about the goal of the task, and introduced its role as an expert assistant in the training phase. Moreover, it moved its hand in a coordinated and synchronized manner with the pendulum movement, to reinforce in the participants the belief that it had direct control over it.
At the beginning of the training session, iCub looked at the participant's face, using a face-tracking module. Then, it assumed a position to get ready to play (bending the torso forward, putting its elbow at 90 degrees angle, with the hands below the table, and looking at the pendulum rod). Then it looked again at participants, inviting them to get ready to play by saying an exhortation, such as "\textit{Ok, I am ready!}" or "\textit{Let's start}".

During the training session, the robot followed the pendulum with its gaze and made prono-supination movements with its forearm, pretending to act on the pendulum. At the end of the session, it invited the subject to take some minutes of rest before starting the next trial.
During baseline and test sessions, iCub remained in its rest position and looked at the subjects performing the task, by alternating its gaze fixation point between the tool and the subject's face.
During the breaks between trials, the robot pretended to take a rest as well, looking around in the room in an exploratory way.
The iCub can show different facial expressions through a set of LED lights representing its eyebrows and lips.
The robot exhibited a happy, friendly face for the whole duration of the experiment, except during the training phase in which he looked at the pendulum with a focused expression (Fig. \ref{Setup}, Right Panel). 
To enhance the impression of animacy, and the pleasantness of the interaction, for the entire duration of the experiment, the eyelids were blinking \cite{Rea_2016}.
Moreover, if the participants left the pendulum fallen, without trying to balance it for more than 5 seconds, iCub turned its gaze to them. Anyway, this behavior rarely occurred (just 3.2\% of the total trials). 
  
A demonstration of the task, experimental setup, and robot behavior can be found in the supplementary downloadable video provided by the authors. This will be available at http://ieeexplore.ieee.org.

\subsection{Questionnaires}

At the beginning of the experiment, participants were asked to fill in a set of questionnaires. 
The questionnaires compiled at this stage by both groups included the Italian version of the TIPI test on participants’ personality \cite{Chiorri_2014} and questions about participants’ previous experience with robots.
From this latter, we found that most of the subjects of the \textit{iCub group} (75\%) had previous experience with robots. However, none of them has ever worked with robots or had experience as robot programmers.

%(9 out of 16 took part in other experiments with robots, 3 of them had already seen robots without interacting with them, and 4 of them had never seen or interacted with robots. None of the participants has ever worked with robots or had experience as robot programmer.

In addition, participants of \emph{iCub group} had to watch a descriptive video\footnote{https://www.youtube.com/watch?v=ZcTwO2dpX8A} of iCub performing several activities and then answer the following questionnaires regarding their perception of it: inclusion of other in self scale (IOS) \cite{Aron_92}, by which it is possible to check how close the person interviewed feel to another subject (in our case the robot); the scales \textit{Anthropomorphism}, \textit{Animacy}, \textit{Likeability}, and \textit{Perceived Intelligence} of the Godspeed questionnaire \cite{Bartneck_2009}; the scales \textit{Mind Experience} and \textit{Mind Agency} of a Mind perception test \cite{Ferrari_2016}, \cite{Gray_2007}; the trust in robots' \textit{Ability} and \textit{Benevolence} \cite{Wang_2016}.
The same items were compiled after the experiment to measure possible changes in participants' perception of the robot. 
When the experiment was concluded, two more questionnaires were given to the participants of both groups: the NASA-TLX workload assessment \cite{NasaTlx} and a short version of the Intrinsic Motivation Inventory (IMI) \cite{Ryan_2000}, comprising 14 items from the sub-scales \textit{Competence}, \textit{Effort/Importance}, and \textit{Interest/Enjoyment}. We randomly distributed the IMI items in the questionnaire and formulated them to fit the specific activity of this experiment. The items were translated into Italian by a professional translator.

Furthermore, we collected participants’ opinions about the experiment through a series of open questions, to get their general feedback about the task and iCub's behaviors. 
After the experiment ended, the experimenter provided subjects with a debriefing and answered their questions and curiosities.
A reward of 15 euros was provided to participants.

\subsection{Data Analysis}

Participants' performance was characterized by using the Performance Index (PI) metric. It was computed by averaging the scores obtained starting from the angular positions of the pendulum.
These scores were derived by evaluating, for each time point, the pendulum angular position in terms of the normalized probability density function of a normal distribution with zero mean and standard deviation of 0.25. 
This was equivalent to giving a score equal to 1 when the pendulum was in equilibrium (the vertical position corresponding to zero degrees), which gradually decreased according to the angle formed with the vertical, and that was equal to zero when the pendulum fell (position = $\pm$ 40 degrees).
In this way, participants who held positions around the vertical for longer were rewarded with higher scores.

%Statistical Analysis
PI was compared among different sessions or between different groups through either paired or independent t-tests, according to the design. When data resulted not following a normal distribution, after a Lilliefors test, the corresponding non-parametric tests were performed. 
The data resulting from questionnaires were also analysed to investigate whether some correlation exists between the perception of the robot iCub and the observed performance by using linear regression. 
The details of each analysis are reported in the results.

\section{RESULTS}
 
\subsection{Performance Evaluation}

\begin{figure}[!tphb]
\centering
\includegraphics[scale = 0.95]{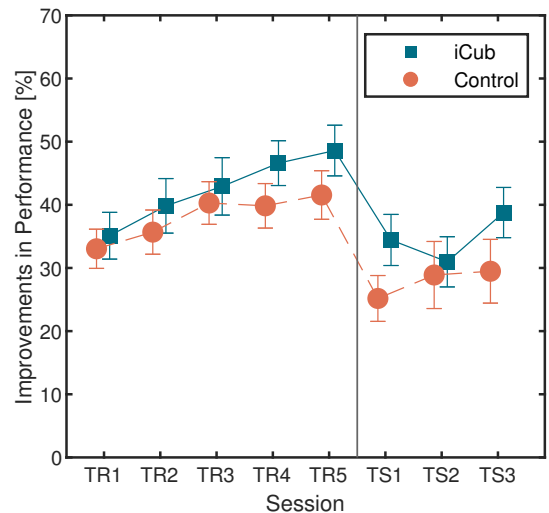}
\caption{Mean and standard error of improvements in Performance Index (PI) with respect to the baseline, for \emph{iCub group} (blue squares) and \emph{Control group} (orange circles) in the five training trials (TR1 - TR5) and the three test trials (TS1 - TS3).}
\label{MeanImprov}
\end{figure}

To test the effect of iCub's involvement on subjects' performance, we compared the PI of the 2 groups in the training and the test sessions, normalized with respect to their initial skill level, i.e. in terms of improvements relative to their baseline performance (Fig. \ref{MeanImprov}).

The baseline performance of the \emph{iCub group} (\textit{M} = 36.5, \textit{SD} = 17.5) and the \emph{Control group} (\textit{M} = 46.1, \textit{SD} = 15) were similar (\textit{t}(27) = 1.59, \textit{p} = .12).

\begin{figure}[!thpb]
\centering
\includegraphics[scale = 0.93]{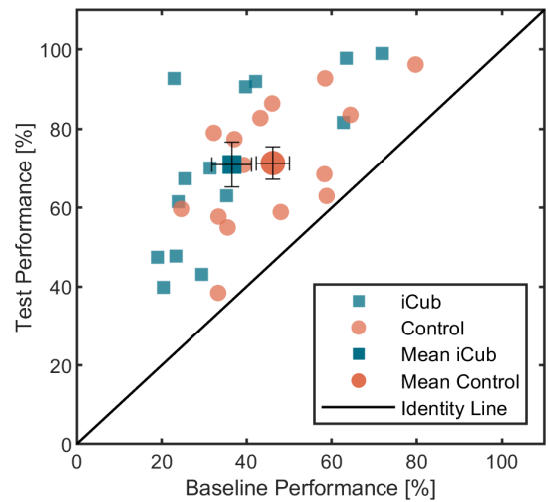}
\caption{Baseline performance versus first test performance of \emph{iCub group} (blue squares) and \emph{Control group} (orange circles), with their respective mean and standard error.}
\label{BLvsTESTperf}
\end{figure}

In the first trial of the training phase, both groups achieved an improvement of their PI of more than 30$\%$.
The performance of the first training trial of both the \emph{iCub group} and the \emph{Control group} (\textit{M} = 71.6, \textit{SD} = 19.4 and \textit{M} = 79.2, \textit{SD} = 8.24 respectively) were significantly higher than their baseline performance (\textit{t}(13) = 9.48, \textit{p} $<$ .001 for \emph{iCub group} and \textit{t}(14) = 10.6, \textit{p} $<$ .001 for \emph{Control group}).
This demonstrated the effectiveness of the implemented assistance, which was designed to provide support and facilitate the task. 
In the successive training trials both groups increased their PI, reaching in the last trial an average improvement of 48.6$\pm$15$\%$ for the \emph{iCub group} and 41.5$\pm$14.9$\%$ for the \emph{Control group}.
In the first trial of test session, the PI improvements dropped down of 14.1$\%$ for the \emph{iCub group} and 16.4$\%$ for the \emph{Control group}. Indeed, the removal of the assistance caused a deterioration of the average performance (\textit{M} = 70.9, \textit{SD} = 21.3 for \emph{iCub group} an \textit{M} = 71.3, \textit{SD} = 16 for \emph{Control group}), which, however, remained significantly higher than the initial baseline values (\textit{t}(13) = 8.51, \textit{p} $<$ .001 for \emph{iCub group} and \textit{t}(14) = 6.93, \textit{p} $<$ .001 for \emph{Control group}).
These results confirm that there has been effective learning in both groups and the ability to generalize the skill also in the absence of assistance. This is also shown in Fig.  \ref{BLvsTESTperf}, where for each participant the performance during test appears higher than the one obtained during baseline. 

In the test session, the average PI improvements among trials of the \emph{iCub group} was equal to 34.7$\pm$3.9$\%$, with a maximum in the last trial and a minimum in the second trial; the mean PI improvements of the control group was equal to 27.8$\pm$2.3$\%$, with a maximum in the last test trial and a minimum in the first one (Fig. \ref{MeanImprov}). 
The average improvements in performance of the 2 groups in the test phase were not statistically different
(\textit{t}(27) = 1.17, \textit{p} = .25).

However, if we consider the individual test trials, the PI improvement of \emph{iCub group} in the first test trial (\textit{M} = 34.4, \textit{SD} = 15.1) is higher than the PI improvement of the \emph{Control group} in the same trial (\textit{M} = 25.2, \textit{SD} = 14), \textit{t}(27) = 1.71, \textit{p} = .049, one-tailed.

The obtained results revealed that the proposed motor training was effective, that it led to generalization, and that the presence of iCub embodying the assistance and showing social behaviors did not interfere with the concentration, but rather led to improvements similar %- sometimes higher - 
to those achieved by the \emph{Control group}.

\subsection{Experience Evaluation} 
The outcomes of the questionnaires filled in the post-experimental phase allowed us to evaluate subjects' overall experience and to compare it between the two groups.

\begin{figure}[!thp]
\centering
\includegraphics[width=3.2in]{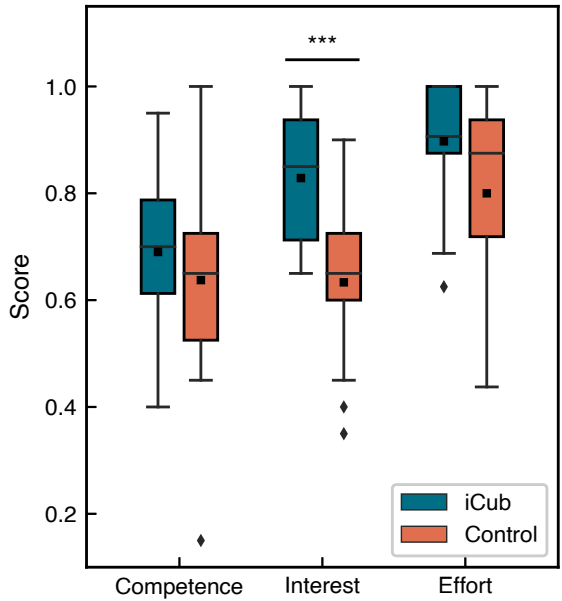}
\caption{Boxplot of normalized scores of IMI questionnaires \textit{Competence}, \textit{Interest} and \textit{Effort} for the \emph{iCub Group} and the \emph{Control group}.} %Lower and upper box boundaries represent 25th and 75th percentiles, respectively; square and line inside box represent mean and median, respectively; lower and upper error lines represent 10th and 90th percentiles, respectively; filled rhombus, data falling outside 10th and 90th percentiles.}
\label{IMI}
\end{figure}

Fig. \ref{IMI} reports the statistics of the normalized scores of the IMI questionnaires, namely \textit{Competence}, \textit{Effort} and \textit{Interest/Enjoyment}; specifically, the latter is considered the self-report measure of intrinsic motivation \cite{Ryan_2000}. 
A Wilcoxon rank sum test indicated that the median score of the sub-scale \textit{Interest} of the \emph{iCub group} (\textit{Mdn} = 0.85), compared to the one of the \emph{Control group} (\textit{Mdn} = 0.65), was significant higher (\textit{Z} = 3.29, \textit{p} $<$ .001).
There was no significant difference for the \textit{Competence} (\textit{Mdn} = 0.70 for \emph{iCub group} and \textit{Mdn} = 0.65 for \emph{Control group}, \textit{Z} = 0.72, \textit{p} = .46) and \textit{Effort} (\textit{Mdn} = 0.91 for \emph{iCub group} and \textit{Mdn} = 0.87 for \emph{Control group}, \textit{Z} = 1.62, \textit{p} = .10) sub-scales, meaning that subjects' perception of their ability and commitment to the task did not change among conditions.

The results reported from the NASA-TLX questionnaire included the answers of 13 participants from the \emph{iCub group} and 14 participants from the \emph{Control group}, as there were problems in the compilation of the forms of 2 participants.
The results showed that the \textit{Total Workload} associated to the task (\textit{Mdn} = 72\%, \textit{IQR} = 18.3\% for \emph{iCub group} and \textit{Mdn} = 65\%, \textit{IQR} = 14.6\% for \emph{Control group}) is not significantly different between the 2 groups (\textit{Z} = 1.21, \textit{p} = .22).

\begin{figure}[thpb]
\centering
\includegraphics[scale = 0.90]{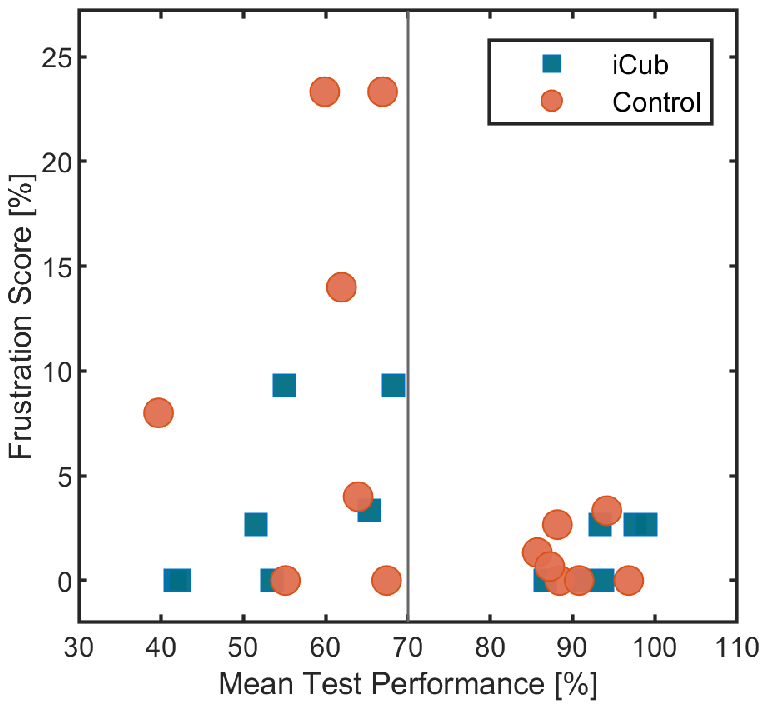}
\caption{Frustration score as a function of mean test performance for \emph{Control group} (orange circles) and \emph{iCub group} (blu squares).}
\label{frustration}
%\end{figure}

%figure regressions
%\begin{figure}[thpb]
\centering
%  \subfigure{\includegraphics[scale = 0.5]{figures/train_antr_sm.eps}}\quadd
%\subfigure{
\includegraphics[scale = 0.95]{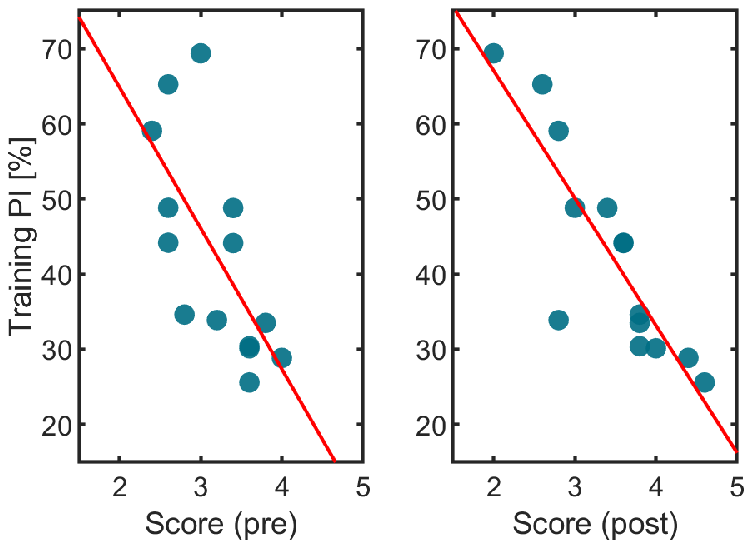}
%}
\caption{Individual PI improvements in the training phase as a function of evaluation of \textit{Anthropomorphism} from the Godspeed questionnaire before and after the experiment.}
\label{regression}
\end{figure}

Considering the scale \textit{frustration}, as shown in Fig. \ref{frustration}, participants who reached high average test performance (above 70$\%$) reported very low scores in both groups. 
In contrast, in the case of test performances below 70$\%$, participants showed more variability in their scores, with a few subjects of the \emph{Control group} manifesting very high frustration levels.
In summary, these results suggest that for both groups the task was similarly challenging, but the presence of iCub influenced the way participants approached the task, making them enjoy more the training experience and potentially mitigating the frustration in case of failure.

% exploring the factors modulating the observed performance
To further explore potential relations between the evaluation of the robot iCub and the performance in the training we conducted some exploratory analyses on the outcomes of the \textit{pre} and \textit{post} questionnaires of the \emph{iCub group} (namely TIPI, \textit{Godspeed}, \textit{Mind Perception}, \textit{Trust} and IOS).
Specifically, we performed linear regressions between questionnaires sub-scales and performance, measured as improvements of PI.

We did not find any significant relationship between our parameters of interest and the TIPI personality traits.

The result for the Godspeed pre-questionnaire indicated that the model containing as predictors \textit{Anthropomorphism} and \textit{Perceived Intelligence} was able to account for 53$\%$ of the variance and that it predicted significantly the mean improvements in training performance, ($R^2$ = 0.53, \textit{F}(2, 11) = 6.12, \textit{p} = .01). The only significant predictor was \textit{Anthropomorphism} (\textit{p} $<$ .04, $\beta$ = -15). 
In other words, participants who ascribed higher anthropomorphic traits to iCub prior to the experiment showed fewer improvements during training.

A strong negative correlation between training performance and \textit{Anthropomorphism} was confirmed further by the outcomes of the post-questionnaire (\textit{Pearson's r}(14) = -0.87, \textit{p} $<$ .001), see Fig. \ref{regression}.

Moreover, the perception of the robot \textit{Animacy} measured in the post-questionnaire was also strongly negatively correlated with average training performance (\textit{Pearson's r}(14) = -0.80, \textit{p} $<$ .001).

Then, we tested whether the performance obtained in the test session correlated with participants' evaluation of the robot. The results showed that the mean improvements in test performance correlates significantly (and negatively) with the \textit{Perceived Intelligence} of the robot (\textit{Pearson's r}(14) = -0.56, \textit{p} = .039), and the sub-scale \textit{Benevolence} of the Trust post-questionnaire (\textit{Pearson's r}(14) = -0.58, \textit{p} = .028). These linear relations indicate that participants who scored worse performances in the test judged the robot to be more intelligent and benevolent.
We did not find any significant relationship between performance and the other questionnaires.

\section{Discussion}

In this study, we wanted to investigate whether and how the presence of a social humanoid robot playing the role of an expert trainer could influence the performance and the experience of naive subjects in the context of motor skills learning. 

We hypothesized that learning with a humanoid robot, showing social behaviors, and embodying the assistance, would bring to more enjoyable training experience, being beneficial in terms of intrinsic motivation and engagement. Also, we evaluated whether this could have repercussions on attention and perceived effort levels, with potential negative effects on performance.
Lastly, we wanted to investigate if, by explicitly presenting the assistance as the contribution of a second agent, we could promote the subjects' ability to generalize the acquired skills when the assistance was removed.
 
To test our hypotheses, we asked participants to learn a complex task, namely to balance an unstable inverted pendulum, in two different experimental conditions: one that involved training with the humanoid robot iCub embodying the physical assistance, and one in which participants had to perform the same training but using the Wristbot alone.
 
Concerning subjects' experience, the results obtained showed that learning in a context of interaction with a humanoid social robot trainer did not have negative effects on attention and effort levels, but it led to greater motivation and more enjoyable training experience. 

As shown by the results of the NASA questionnaire, the subjects of the two groups presented very similar values of cognitive workload showing that the way they perceived the task and the effort made to accomplish it did not change among conditions. Of interest, when asked to rate their feeling of frustration, only a few participants in the \emph{Control group} provided high frustration scores.
In addition to this, as reported through the IMI questionnaire, the involvement of iCub had a very strong effect on the subjects' intrinsic motivation, i.e. the interest and enjoyment they felt while learning a very challenging and arduous task. 
Since most of the participants were familiar with robots, we believe that this increase in interest and enjoyment is not attributable to the effect of interacting with novel technology.
This is an implication that should not be underestimated, especially when considering rehabilitation contexts. Motivation is key to establish adherence to a therapeutic or learning path and to promote behavioral changes. The presence of a humanoid social agent could change significantly the way the patients approach the training, increasing their commitment and willingness to train together with their chance to succeed.

%% parte sulla linear regression
The outcomes of the questionnaires about iCub perception showed that subjects who ascribed higher anthropomorphic traits to iCub prior to the experiment showed fewer improvements during training (although maintaining a relatively good absolute performance). 
This suggests that the tendency to anthropomorphize the robot's behavior might be associated to a different impact of the humanoid's presence on the learning, e.g. inducing participants to focus relatively more on the robot than on the task. 

Moreover, the negative correlation between the tendency to judge the robot as anthropomorphic and the improvements in performance was confirmed also in the post-questionnaires. Similarly, people who achieved lower improvements during training, attributed to robot higher \textit{Animacy}.

Again, a possible explanation is that those who achieved higher performance improvements were focused mostly on the pendulum and the correct execution of the task, being less intrigued by the humanoid’s behavior.
This result could be explained also by the fact that people with lower performance felt more the assistance contribution during training, being potentially induced to increase their ratings about iCub social properties. 

Finally, we found that the robot was judged more benevolent and intelligent by participants who exhibited worse performance improvements in the test phase. We speculate that naive subjects who encountered more difficulties in the test phase attributed their previous good outcomes in the training to the assistance of a benevolent and expert robot.

Considering skills improvements, the obtained results bring us to conclude that the presence of the robot did not have negative effects on the performance of the subjects; on the contrary, it seemed to lead to similar improvements and to facilitate the process of generalization. We found indeed a tendency to show higher performance improvements in the \emph{iCub group} as soon as the assistance was removed (i.e. in the first trial of test session). 
This result leads us to speculate about how humans can generalize the acquired model of a certain task when this is carried out in a situation of social interaction. When training is facilitated by assistance, the task representation created by the naive learner may be misled, including the assistance as part of the task dynamic rather than as a transitional contribution. We suggest that learning in an interaction context favors the creation of separated representations of the task and the assistance, with the latter being associated with the intervention of the other agent. Although more evidence is needed, our results seem to encourage this theory, underlining the need for further investigations. 

It is worth noting that the iCub robot showed social behavior, but this was not modulated by the performance of the participant, nor was it responsive to participants' reactions. The robot did not encourage or congratulated and did not even change its facial expression according to good or bad human performance. Even so, the mere intervention of a seemingly social humanoid assistant had an impact on participants' training. This expands to the context of learning previous results obtained in the perceptual domain, where the mere exhibition of social behavior by a humanoid robot had a significant impact on the human partner's perception \cite{Mazzola_2020}.
Then, we expect that endowing the robot with the possibility to adapt its social response to the partner's status will increase dramatically the effect of the robot tutoring on participants' experience.

%%limiti
The limited interactivity of the robot in our experiment was due to two main factors: firstly, as explained before, we made this choice to keep the two experimental conditions as comparable as possible; secondly, the participants stated that their interaction with the robot was limited by the fact that they needed complete concentration during the accomplishment of the task, and so their attention was focused mostly on the pendulum rather than on the humanoid robot.    
However, during the resting phases, most of them tried to establish an interaction with iCub and, during the test phase, when the assistance was no longer present, many turned to iCub to ask it to intervene again in the task or to note the improvements achieved.

As a final consideration, we can state that during the training of demanding tasks with high cognitive load, where the attention is turned to a specific external stimulus, the interaction between the naive subject and trainer robot is very moderate. This phase of interaction must, therefore, exploit specific and effective channels of communication, potentially different from those used in absence of a concurrent task (e.g., during pauses) where more subtle communication signals can be effective. 

With the novel experimental setup presented in this study, where it is possible to record and send both haptic signals, by using the robotic handle Wristbot, and social cues, through the humanoid robot iCub, we want in the future to further study the relative contribution of these different communication channels to the context of human-robot interaction for motor skills learning.

Our long-term goal is to provide robots with an adaptive assistive architecture, which will allow us to get closer to robots that are not just assistive devices but rather assistive partners, able to guide humans in both short-term and long-term processes of skills learning and recovery, and to adapt to their needs through a customized interaction. Adaptation and learning need to occur both on the side of the robot and the human.
In particular, robots should adapt and learn how to improve engagement or dose the level of assistance in training activities while interacting with different persons. 

\addtolength{\textheight}{-12cm}   % This command serves to balance the column lengths

%%%%%%%%%%%%%%%%%%%%%%%%%%%%%%%%%%%%%%%%%%%%%%%%%%%%%%%%%%%%%%%%%%%%%%%%%%%%%%%%

%%%%%%%%%%%%%%%%%%%%%%%%%%%%%%%%%%%%%%%%%%%%%%%%%%%%%%%%%%%%%%%%%%%%%%%%%%%%%%%%

%%%%%%%%%%%%%%%%%%%%%%%%%%%%%%%%%%%%%%%%%%%%%%%%%%%%%%%%%%%%%%%%%%%%%%%%%%%%%%%%
%\section*{APPENDIX}


%\section*{ACKNOWLEDGMENT}

%%%%%%%%%%%%%%%%%%%%%%%%%%%%%%%%%%%%%%%%%%%%%%%%%%%%%%%%%%%%%%%%%%%%%%%%%%%%%%%%

\end{document}